\def\year{2021}\relax
\documentclass[letterpaper]{article} 
\usepackage{aaai21}  
\usepackage{times}  
\usepackage{helvet} 
\usepackage{courier}  
\usepackage[hyphens]{url}  
\usepackage{graphicx} 
\usepackage{natbib}  
\usepackage{caption} 
\usepackage{multirow}
\usepackage{amsmath}
\usepackage{booktabs}
\usepackage{arydshln}
\usepackage{enumerate}
\usepackage{amsfonts}
\usepackage{amssymb}
\usepackage{pgfplots}
\usepackage[switch]{lineno}  %
\title{Graph-to-Sequence Neural Machine Translation }
\author {
    Sufeng Duan \textsuperscript{\rm 1,2,3}, 
Hai Zhao \textsuperscript{\rm 1,2,3 \footnote{Corresponding author. This paper was partially supported by National Key Research and Development Program of China (No. 2017YFB0304100) and Key Projects of National Natural Science Foundation of China (No. U1836222 and No. 61733011).}}, Rui Wang \textsuperscript{\rm 4}  \\
}
\affiliations{
    \textsuperscript{\rm 1}Department of Computer Science and Engineering, Shanghai Jiao Tong University \\
	 \textsuperscript{\rm 2}Key Laboratory of Shanghai Education Commission for Intelligent Interaction \\ and Cognitive Engineering, Shanghai Jiao Tong University, Shanghai, China\\
	 \textsuperscript{\rm 3}MoE Key Lab of Artificial Intelligence, AI Institute, Shanghai Jiao Tong University \\
	 \textsuperscript{\rm 4}National Institute of Information and Communications Technology (NICT), Kyoto, Japan\\
{\tt 1140339019dsf@sjtu.edu.cn, zhaohai@cs.sjtu.edu.cn, wangrui@nict.go.jp}
}

\begin{document}


\maketitle

\begin{abstract}
Neural machine translation (NMT) usually works in a seq2seq learning way by viewing either source or target sentence as a linear sequence of words, which can be regarded as a special case of graph, taking words in the sequence as nodes and relationships between words as edges. In the light of the current NMT models more or less capture graph information among the sequence in a latent way, we present a graph-to-sequence model facilitating explicit graph information capturing. In detail, we propose a graph-based SAN-based NMT model called Graph-Transformer by capturing information of subgraphs of different orders in every layers. Subgraphs are put into different groups according to their orders, and every group of subgraphs respectively reflect different levels of dependency between words. For fusing subgraph representations, we empirically explore three methods which weight different groups of subgraphs of different orders. Results of experiments on WMT14 English-German and IWSLT14 German-English show that our method can effectively boost the Transformer with an improvement of 1.1 BLEU points on WMT14 English-German dataset and 1.0 BLEU points on IWSLT14 German-English dataset.
\end{abstract}

\section{Introduction}
Encoder-decoder architecture, which uses an encoder to create a representation of source sequence and a decoder to predict target sequence, have been established as state of the art approaches in neural machine translation (NMT)  \cite{kalchbrenner2013recurrent, DBLP:conf/nips/SutskeverVL14, DBLP:journals/corr/BahdanauCB14}. Recurrent neural network based (RNN-based) model \cite{DBLP:conf/nips/SutskeverVL14, DBLP:journals/corr/BahdanauCB14,DBLP:conf/cncl/WuZ18}, convolutional neural network \cite{DBLP:conf/icml/GehringAGYD17} model and self-attention network based (SAN-based) model \cite{DBLP:conf/nips/VaswaniSPUJGKP17} are representative encoder-decoder models, and most of NMT models are variants or combination of these three. NMT models based on encoder-decoder architecture are similar in
 some aspects, such as stack of layers having the same structure.

Stack of layers increases the complexity of model to approximate nonlinear function. Viewing all layers as one function, every single layer captures different information from input. Looking into every single NMT model such as RNN-based model or SAN-based model, models always try to make representation of one word containing information of whole sentence in every layer. However, empirically, one layer alone cannot result in satisfactory result. 

It is common to regard sentence in NMT model as a directed complete simple graph, which views words as nodes and relationships between words as edges. However, this perspective only focuses on relationship between words, while ignoring other information, such as relationship between phrases or relationship between different fragments of sentences. As a result, structure of simple graph cannot fully reflect all information. 

To overcome the shortcomings of simple graph, we view sentence as a multigraph $G$ in SAN-based model. In multigraph $G$, multiple edges exist between two nodes. Edge connects not only nodes but also subgraphs of $G$ which reflects relationship between different fragments of sentences more than relationship of word-pair. Encoding is also regarded as a process of generating a multigraph to approximate $G$ infinitely. Compared with simple graph, multigraph can explain th essence of encoding more comprehensively, and explain relationship between words in a more general way.

One layer in NMT model can capture the incremental information automatically compared with its previous layer. Fusion of the previous and incremental information makes representation more rich and thus benefits translation. From the perspective of multigraph, incremental information can be described as a set of higher-order subgraphs generated by this layer. Even though the current NMT models can capture information of subgraphs of different orders, fusing them into a representation with a fixed weight makes the model difficulty to pay more attention on really salient part. 

To solve this problem, we propose a graph-based SAN empowered Graph-Transformer by enhancing the ability of capturing subgraph information over the current NMT models. First of all, we generally define a \textbf{full representation} as the fusing result of all concerned subgraph representations. Then let the representation of one layer split into two parts, \textbf{previous representation} and \textbf{incremental representation}. The previous representation reflects full representation from previous layer, and the incremental representation reflects new information generated in this layer. Based on this, the encoding process is modified to adapt to such representation division. We split the original self-attention into three independent parts to generate incremental representation. Our method accommodates subgraphs of different orders into different parts of incremental representation, and reduces the information redundancy. To fuse the full representation, We consider three fusing strategies in terms of different weighting schemes so that let the model focus on important parts of representation.

In experiments on WMT14 English-to-German (En-De) and IWSLT14 German-to-English (De-En), results of experiments prove our model can improve performance of translation with a few parameters increasing. Our model achieves a performance outperforming the Transformer with an improvement of 1.1 BLEU points in En-De and 1.0 BLEU points in De-En.

\section{Background}

Transformer \cite{DBLP:conf/nips/VaswaniSPUJGKP17} is state-of-the-art NMT model empowered by self-attention networks (SANs) \cite{DBLP:conf/iclr/LinFSYXZB17}, in which an encoder consists of one self-attention layer and a position-wise feed-forward layer,  decoder contains one self-attention layer, one encoder-decoder attention layer, and one position-wise feed-forward layer. SAN-based NMT model uses residual connections around the sublayers followed by a layer normalization layer. 

The encoder reads an input sentence, which is a word sequence $x=\{x_1,...x_{T_x}\}$, and encodes it as a context vector $c$. Decoder is trained to predict the next word given the context vector generated by encoder and all previously predicted words $\{y_1,...,y_{t-1}\}$. The decoder defines a probability over the translation $y$ by decomposing the joint probability into the ordered conditionals,
\begin{equation}
    \label{eq:decoder_prob}
    p(y) = \prod_{t=1}^{T_y} p(y_t \mid \left\{ y_1, \cdots, y_{t-1} \right\}, c).
\end{equation}

Scaled dot-product attention is the key component in Transformer. The input of attention contains queries ($Q$), keys ($K$), and values ($V$) of input sequences. The attention is generated using queries and keys like Equation (\ref{orisof}),

\begin{equation}
\label{orisof}
Attention(Q, K, V) = {\rm softmax}(   \frac{Q{K^T}}{\sqrt{d_k}})V.
\end{equation}
Different from RNN-based models which process words/subwords one by one, dot-product attention allows Transformer to generate the representation in parallel.

\citet{DBLP:conf/nips/VaswaniSPUJGKP17}  also propose multi-head attention which generates representation of sentence by dividing queries, keys, and values to different heads and gets representative information from different subspaces. 

Quality of encoding sentence and generating representation can influence performance of NMT model significantly. RNN-based and SAN-based models use different mechanisms to implement encoding, thus show different natures for the resulted representation. RNN-based model is good at capturing localness information and not good at parallelization and long-range dependency capturing, while SAN-based model is better at capturing long-range dependencies with excellent parallelization.

\section{Graph-aware Representation}
A sentence can be viewed as a sequence of words. To get a generalized formalism over the sentence structure, we may view a sentence as a multigraph which views words as nodes and relationships between words as edges. This section introduces our proposed perspective for graph-aware representation of source sentence.

\subsection{Encoding of Input Sentence}
General speaking, encoding of input sentence is a process to transfer a sequence of words to a sequence of vectors which is composed of number. During encoding, model is treated as a stable function independent of data. Representation generated by model only reflects information of input sentence.

\subsection{View Sentence as a Multigraph}
\label{subsection:multigraph}
We define a directed multigraph $G=(V,E,SN,TN)$ instead of directed simple graph over a sentence $S=\{s_1,...,s_n\}$, in which nodes $V=\{v_1,...,v_n\}$ reflect words of $S$, edges $E=\{e_1,...,e_m\}$ reflect relationship between words of $S$, $SN=\{sn_1,...,sn_m|sn_j \in V, 1<j\le m\}$ is the set of source node of each edge in $E$ and $TN=\{tn_1,...,tn_m|tn_j \in V, 1 < j \le m\}$ is the set of target node of each edge in $E$. Node in $G$ can access other nodes in one step. Put all words of $S$ in $G$, $G$ can also be a representation of $S$. Information captured from $S$ is splited into two parts, (1) \textbf{Word information}, which are contained in $V$ and reflects word, (2) \textbf{Relationship information}, which are contained in $E$ and reflects relationship of word-pairs.

Obviously, one word information is independent of other words, and model cannot enrich word information, which means that word information has been determined before encoding and cannot increase during encoding. The total amount of relationship information is limited because the number of words is limited. So the amount of information which can be captured from $S$ is limited, which means that $G$ is certain. But $G$ is still unknown for model. So generating representation of input sentence can be regarded as a process of generating a multigraph to approximate $G$ infinitely.

\subsection{Edge and Subgraph}
\label{subsection:subgraph}
In this section, we introduce edge and subgraph from a perspective on generating representation during encoding.

A subgraph of $G$ is a graph whose vertex set is a subset of $V$, and whose edge set is a subset of $E$. We define $Sub_G=\{sub^G_1,...,sub^G_p\}$ as the set of all subgraphs of $G$. Subgraph can be defined as $sub^G_j=(V^G_j,E^G_j,SN^G_j,TN^G_j)$. Order of subgraph $sub^G_j$, which is equal to $|V^G_j|$, is the number of nodes in it.  The simplest subgraph has one node and no edge, and order of it is 1.

Edges reflect not only relationship between words, but also relationship between subgraphs.  Given one node-pair, several edges are generated because nodes may belong to different subgraph-pairs. $p(v_i, v_i \in V^G_k|v_j, v_j \in V^G_h)$ is the conditional probability to present one relationship between $v_i$ and $v_j$. It indicates that edge $e_j$ is determined by four variables, (1) source node of edge $sn_j$, (2) target node of edge $tn_j$, (3) subgraph $sub^G_k$ in which $sn_j \in V^G_k$, (4) subgraph $sub^G_h$ in which $tn_j \in V^G_h$. We can use quaternion $(sn_j, tn_j, sub^G_k, sub^G_h)$ to present edge $e_j$.  If we only focus on source and target node, we can use $(sn_j \rightarrow tn_j)$ for $e_j$.

\begin{figure}
\centering
\includegraphics[scale=0.18]{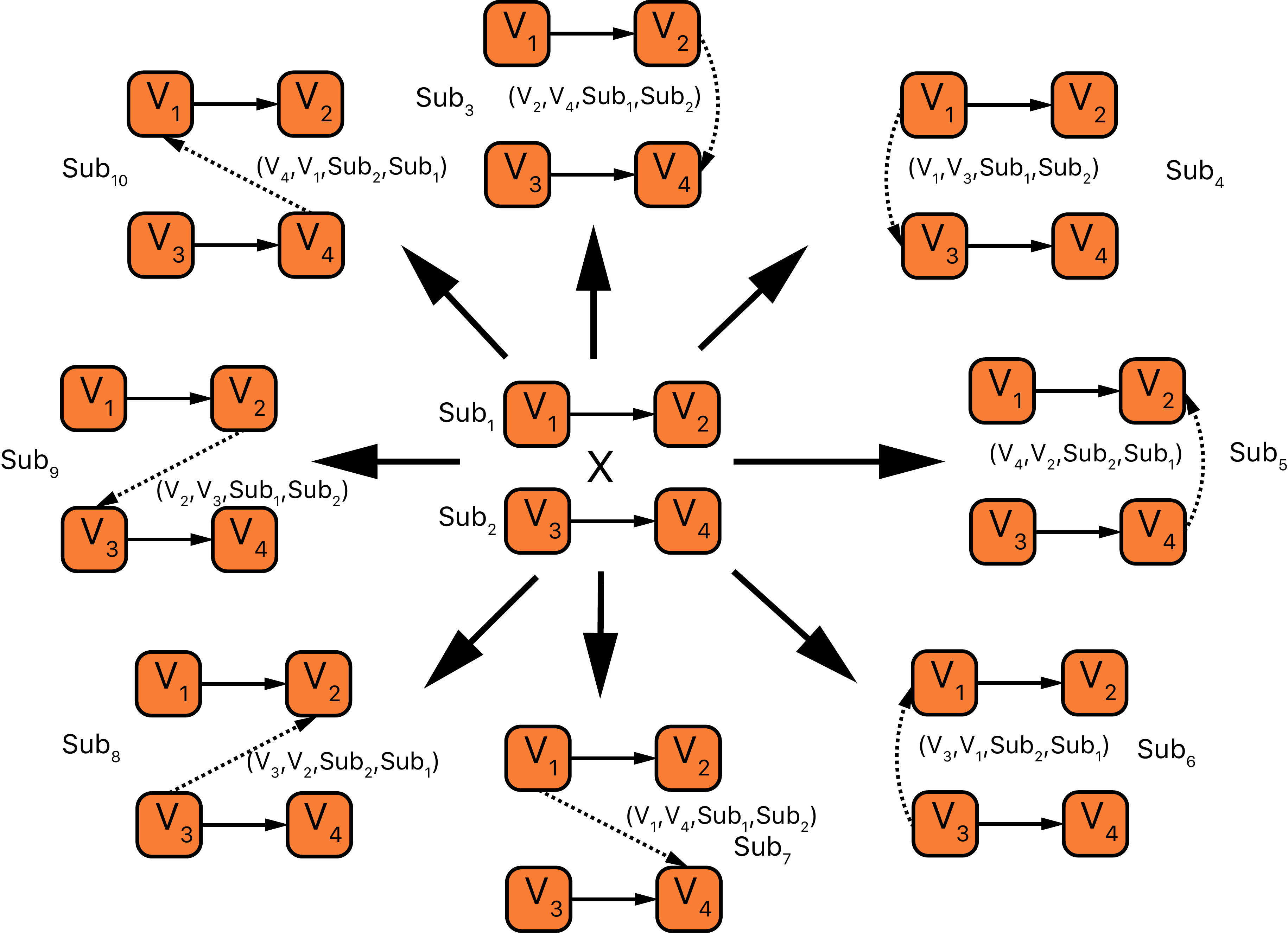}
    \caption{Example of generating novel subgraphs.}
    \label{fig:captures}
\end{figure}

Every edge reflects relationship of one node-pair (subgraph-pair) and connects two subgraphs, which generates one novel subgraph with all nodes and edges from old subgraphs. As a result, order of novel subgraph is larger or equal to order of given subgraphs, and subgraphs of lower order should be generated earlier. Note that the edge connecting two subgraphs must be generated after generating two subgraphs, which means this edge cannot be generated before generating edges in these two subgraphs. Therefore, the order of generaing edges cannot change, which also means that the order of generating subgraphs cannot change. Figure \ref{fig:captures} is an example of generating novel subgraphs.

\subsection{Multigraph in SAN-based Model}
Regarding representation as a multigraph, subgraph can be regarded as the main body of representation. Because we can only generating a multigraph to approximate $G$, representation which we generate is only a subgraph of $G$, or a subset of $Sub_G$. In NMT model, representation is presented as a one-dimensional matrix, which means to represent a subgraph, we may use a matrix with the same shape as representation. Given representation for a set of subgraph, representation-pair can be expanded into a set of subgraph-pair. Given two representation $r_a$ and $r_b$, $\{sub_1^a,...sub_n^a\}$ are $n$ subgraphs contained in $r_a$ and $\{sub_1^b,...sub_m^b\}$ are $m$ subgraphs contained in $r_b$ as Equation (\ref{equation:representation}) shown. 
\begin{equation}
\begin{aligned}
r_a &=\sum_i^n sub_i^a, r_b=\sum_i^m sub_i^b.
\end{aligned}
\label{equation:representation}
\end{equation}

In SAN-based model, self-attention has to get an attention matrix $\mathcal{M}$ using query and key according to Equation (\ref{orisof}). The value in $\mathcal{M}$ reflects relationship (edge $e$) of one word-pair. In the first layer, all information of word-pairs can be found in $\mathcal{M}$. However, in later layers, there is supposed to be more relationship information contained in this matrix after multi-layer learning. In other words, the value in later layer does not only reflect relationship of word-pair. In fact, it also potentially reflects the relationship between word representations which presents subgraphs, same as the edge we discussed above.

In the $i$-th layer of SAN-based model, representation of word $s_m$ of sentence $S$ generated by this layer is $r_m^i$, and representation of word $s_m$ is $r_m^{i-1}$. In the first layer, this representation is generated by word embedding. According to Equation (\ref{orisof}), $r_m^i$ can be calculated by

\begin{equation}
   \begin{aligned}
      r_m^i=\sum^n_jva_{mj}^ir_j^{i-1},
    \end{aligned}
    \label{eq:rmi}
\end{equation}

\noindent where $va_{mj}^i$ is value which is in row $m$ and column $j$ of the matrix $\mathcal{M}_i$ calculated by query and key in $i$-th layer. Note that $va_{mj}^i$ is different from $va_{jm}^i$. Equation (\ref{orisof}) shows that self-attention extracts relationship between every input representation-pair and combines two representations into one new. Because $va_{mj}^i$ is calculated by $r_m^{i-1}$ and $r_j^{i-1}$, $va_{mj}^j$ can reflect fusion and relationship of $r_m^{i-1}$ and $r_j^{i-1}$.

To calculate matrix $\mathcal{M}_i$, if $r_a$ is query and $r_b$ is key,
\begin{equation}
    \begin{aligned}
        r_a \cdot (r_b)^T &=(\sum_i^n sub_i^a) \cdot (\sum_i^m sub_i^b)^T \\
        &=\sum_i^n \sum_j^m sub_i^a \cdot (sub_j^b)^T
    \end{aligned}
    \label{eq:submatrixs}
\end{equation}
Note that relationship between $sub_i^a$ and $sub_j^b$ is calculated by $sub_i^a \cdot (sub_j^b)^T$. Equation (\ref{eq:submatrixs}) shows that $va_{mj}^i$ in $\mathcal{M}$ is a sum of all relationship of subgraph-pairs. However, self-attention can generate relationship and fail to weight them. These relationship are added to $va_{mj}$ with the same weight.

\begin{figure}
    \centering
    \includegraphics[scale=0.15]{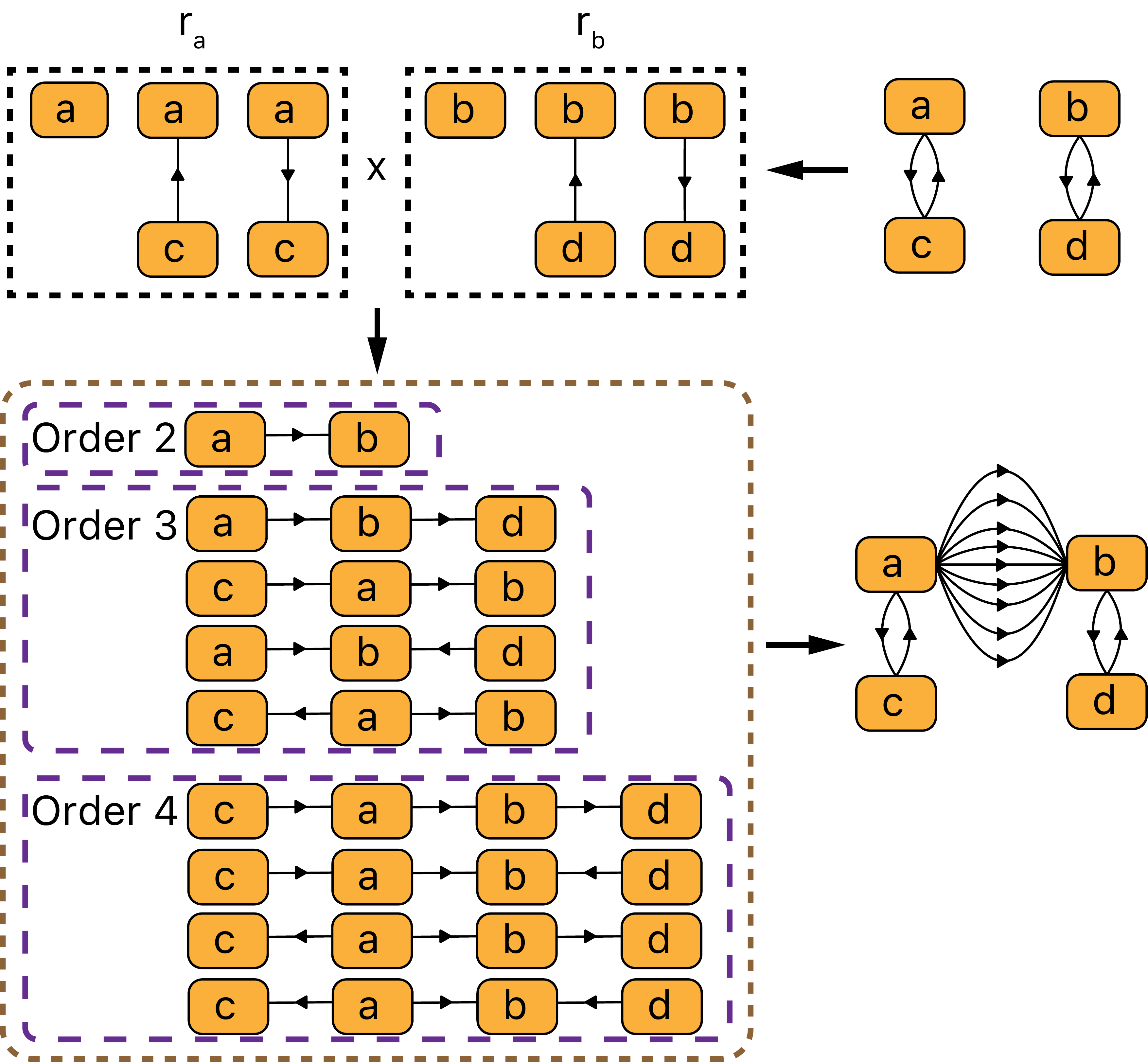}
    \caption{An example of subgraph layers captured. Notes that there are several edges between two nodes.}
    \label{fig:subgraph_label}
\end{figure}

In the first layer, all relationships of word-pairs which involve word $s_m$ are generated by self-attention and contained in $r_m^1$. Given $s_m$ and $s_n$, $r_m^1$ contains word information of $s_m$ and $s_n$, and edge $(v_m \rightarrow v_n)$. It means that a subgraph which has $s_m$ and $s_n$ as nodes and edges between them can be rebuilt by using $r_m^1$. So $r_m^1$ can be viewed as a set of subgraphs whose orders are not more than 2. Likewise, other layers can expand representation(-pair) from previous layers into a set of subgraph(-pair). With an edge connecting two subgraph, layer can generate a subgraph with all nodes and edges from the two subgraphs. In SAN-based model, this new edge is calculated by self-attention once. As a result, each layer encodes a set of subgraphs into representations, and the order of subgraps generated by $i$-th layer is from $2^{i-1}$ to $2^i$. Maximum order of subgraph is based on number of layers in SAN-based model. The more layers model have, the more complex subgraphs the model can learn. Figure \ref{fig:subgraph_label} is an example of generating subgraphs.

RNN-based and SAN-based models can continuously approximate the representation of whole sequence in the process of generating subgraph representations. 
Theoretically, a SAN-based model with $n$ layers can generate representations for all subgraphs whose orders are not more than $2^n$. It means that all possible subgraphs will be 
generated by SAN-based model if sentence is not longer than $2^n$. It does not mean that such subgraph representation is only an inconsequential information. Actually, subgraph allows model to focus on salient part of input sequence such as phrases.

However, nether RNN-based model nor SAN-based model can accurately obtain information of the whole sentence and all the concerned subgraphs. Mechanism of forgetting makes RNN-based model generate the subgraph representation incomplete and it is worse that there is no explicit way to get known any subgraphs captured or ignored. For SAN-based model, the largest order of subgraph is limited by the number of layers. In most cases, the  model can only use part of all subgraphs to make a prediction.

\subsection{Weight Subgraph}
Despite of the limited model capability, SAN-based model may pay more attention to lower-order subgraphs naturally. This is because that models always learn or predict repeatedly and incrementally and subgraphs of low order are thus captured repeatedly. The earlier subgraph is generated, the larger weight it can get. Such a procedure increasess the weight of low-order subgraphs in a latent way.

Putting multiple subgraphs in one vector makes model difficult to distinguish them and extract important subgraphs from vector. In the meantime, illegal subgraphs, such as $a\rightarrow b\rightarrow a$, may cause serious performance loss.

Instead of weighting every detailed subgraphs, it is more practical to group subgraphs by order and weight every group. In every layer, we control subgraph generating and grouping, so that the model can focus on some specific groups of subgraphs. That is the idea this work is based on.

\section{Graph-Transformer}
\begin{figure}
    \centering
    \includegraphics[scale=0.23]{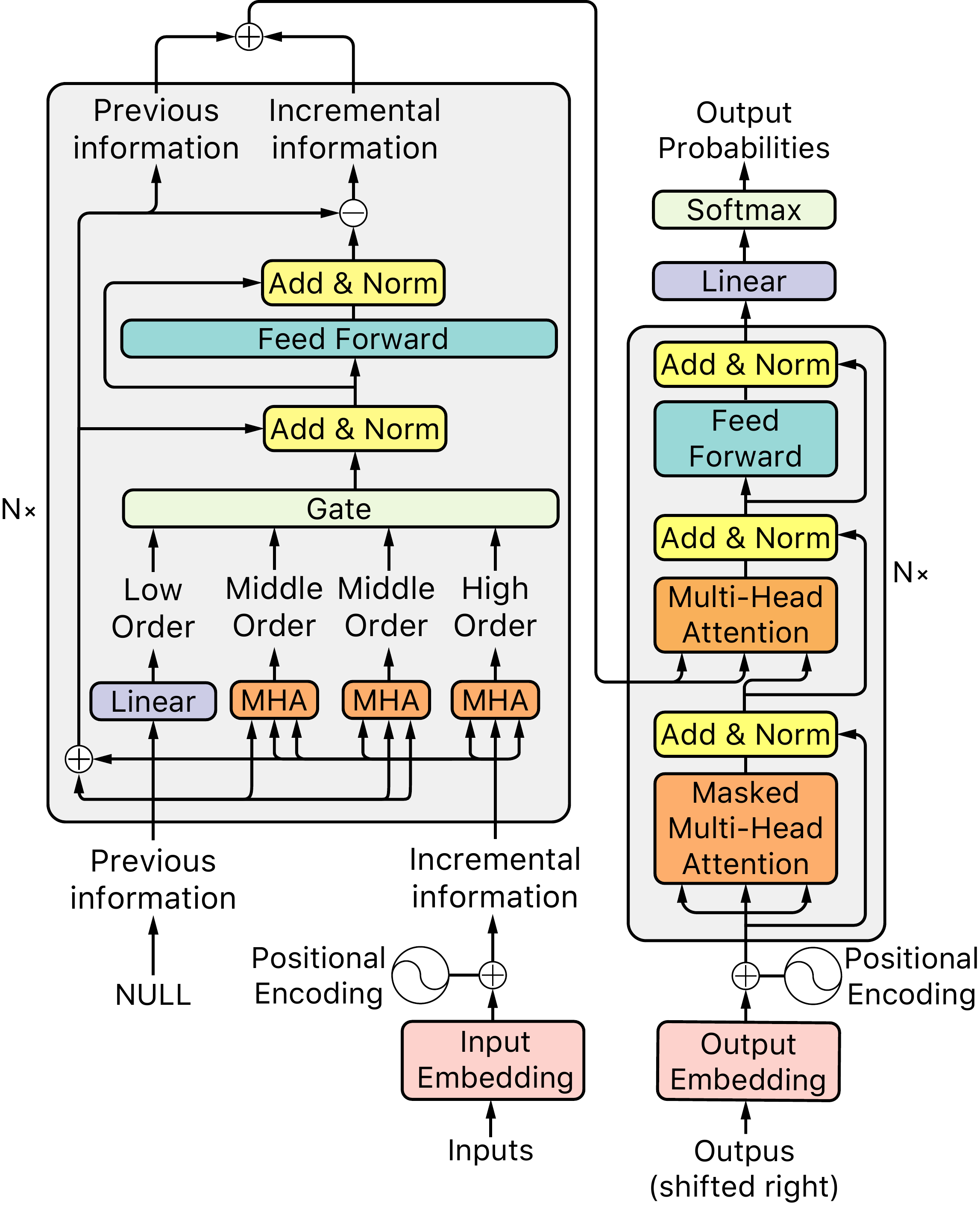}
    \caption{Architecture of model. MHA is short for Multi-head Attention.}
    \label{fig:models_label}
\end{figure}

Figure \ref{fig:models_label} is the architecture of our proposed graph-Transformer. First, we generally define a \textbf{full representation} as the fusing results of all concerned subgraph representations. To capture subgraphs and group them, we split full representation of layer into \textbf{previous representation} and \textbf{incremental representation}. The previous representation is sum of two input representations of one layer, which is also full representation from previous layer. Incremental representation is new information generated by one layer. Self-attention is split into three parts to generate incremental representation. One part of self-attention only focus on one part of representation, which reflects one group of subgraphs. After encoding, the sum of previous and incremental representations will be viewed as the final representation.

\subsection{Self-Attention Group}
We put all subgraphs into four groups, and self-attention process three groups. Different from the original Transformer which uses same representation as query, key and value, every part of self-attention in our encoder has different query, key and value. Three parts of self-attention capture subgraph of different orders and self-attention can capture three groups of subgraphs of higher order than input representation of layer.  There are three levels for the subgraph order.

$\bullet$\textbf{High order.} One group of subgraphs belong to high order. The first part of self-attention is used to process this group, which uses input incremental representation as its query, key and value. In the $i$-th layer, the order of subgraph is in the range of $2^{n-1}$ to $2^n$.

$\bullet$\textbf{Middle order.} Two groups of subgraphs belong to middle order and other two part of self-attention are used to process them. The second part of self-attention uses input incremental representation as its query and input previous representation as its key and value. The third part of self-attention uses input previous representation as its query and input incremental representation as its key and value. In the $i$-th layer, the order of subgraph is in the range of $2^{n-2}$ to $2^{n-1}$. Note that the second part and the third part of self-attention captures subgraph with the same order.

$\bullet$\textbf{Low order.} For one layer, previous layer has generated enough subgraph of low order and it is no need to capture subgraph of low order again. Therefore subgraphs of low order come from previous layer and there is no self-attention to process it. The subgraph order is in the range of 1 to $2^{n-2}$.

In our model, three parts of self-attention use two kinds of representations as query, key and value, which means six vectors are enough.
 Shared vector of query, key and value can reduce number of parameters and thus avoid overfitting while it is difficult to train such a model because different sets of subgraphs may require different vector spaces.

Note that there is always redundancy of subgraph with the same order, but it is necessary for model to extract salient information from the redundancy. 
Capturing subgraph by each invidual parts may reduce redundancy because one subgraph will be generated only once. 

To alleviate the redundancy of subgraph representation, we can drop the dimension of model in self-attention. 
To keep the least effect over the performance, we can reduce dimension of model to half of original dimension.

\subsection{Fusion of Subgraphs}

We call a fusion representation of four groups of subgraphs as \textbf{full representation}. \textbf{Previous representation} of one layer is the full representation generated by previous layer. \textbf{Incremental representation} is defined as the difference between full representation and previous representation. 

To get the full representation, we consider three fusing strategies. Calculating the sum is the most simple one to implement. However, it depends on the quality of representation. In this case, model gives four groups equal weights which is similar as the original Transformer.

Three parts of representation generated by self-attention are not weight by model. Previous representation is a full representation generated by previous layer and can be viewed as one group. So we can put new generated representations in one group. To weight two groups, we use a gate to calculate their importance and merge them.

\begin{equation}
\begin{aligned}
w={\rm Sigmoid}(i_h+i_m+i_l),\\
r_f=(i_h+i_m) \dot w + i_l\dot(1- w)
\end{aligned}
\end{equation}

\noindent where $i_h$, $i_m$ and $i_l$ are subgraph of high order, middle order and low order. With gate to assign weight, model can explicitly distinguish new and old subgraph and pay attention on important group of subgraph. Disadvantage of this method is that the model still cannot distinguish subgraph of high and middle order. We call this method \textbf{weight-gate}.

Self-attention is also a choice by generating a matrix of weight which stands for relationship between words. \cite{wang-etal-2018-multi-layer} design a fusion function based on the self-attention model with hops for fusion of representation of different layers. Similar as \cite{wang-etal-2018-multi-layer}, we deal with several representations of subgraph, and concatenating four representations to form a new sequence $R$. Self-attention will calculate the matrix of relationship between different representations and update every unit in $R$.

\begin{equation}
\begin{aligned}
R_f
&= \frac{softmax(\frac{R_q{R_k^T}}{\sqrt{d_k}})}{4}R_v
\end{aligned}
\end{equation}
where $R_f$ is the representation sequence, $R_q$, $R_k$ and $R_v$ are vector of query, key and value, $d_k$ is the dimension of model. This method can capture relationships between representations and weight them. Weight of one group will be larger if it is more important than others. To make sum of weight equal to 1, representation is divided by 4. We call this method \textbf{self-gate}.

An ideal result is that self-gate can weight every group of subgraph and generate a better representation compared with the original Transformer. However, according to the discussion above, redundancy of subgraph may be generated by self-attention and become a noise which influences performance. Self-gate may also generate subgraphs of higher order which is similar as making model deeper and makes model difficult to be trained.

\section{Experiments}
\subsection{Datasets}
We evaluate our model on two translation tasks, IWSLT14 German-English (De-En) and WMT14 English-German (En-De).

\noindent\textbf{IWSLT14 German-English} IWSLT14 De-En dataset contains 153K training sentence pairs. We use 7K data from the training set as validation set and use the combination of dev2010, dev2012, tst2010, tst2011 and tst2012 as test set with 7K sentences which are preprocessed by script\footnote{https://github.com/pytorch/fairseq/blob/master/examples/\\translation/prepare-iwslt14.sh}. BPE algorithm is used to process words into subwords, and number of subword tokens is 10K. 

\noindent\textbf{WMT14 English-German} We use the WMT14 En-De dataset with 4.5M sentence pairs for training. We use the combination of newstest2012 and newstest2013 as validation set and newstest2014 as test set which are preprocessed by script\footnote{https://github.com/pytorch/fairseq/blob/master/examples/\\translation/prepare-wmt14en2de.sh}. The  sentences longer than 250 are removed from the training dataset. Dataset is segmented by BPE so that number of subwords in the shared vocabulary is 40K.

\begin{table}[htb]

\newcommand{\tabincell}[2]{\begin{tabular}{@{}#1@{}}#2\end{tabular}}
\begin{center}
\begin{tabular}{lccccc}
\toprule
\multirow{1}{*}{Parameter} & \multicolumn{1}{c}{DE-EN}&\multicolumn{1}{c}{EN-DE} &  \\

\hline
\tabincell{c}{Layers}&6&6 \\
\tabincell{c}{Dimension } &512&512 \\ 

\tabincell{c}{Head} &\tabincell{c}{4}&\tabincell{c}{8} \\
\tabincell{c}{FF} &1024&2048 \\
\tabincell{c}{Dropout} &0.3&0.1 \\
\bottomrule
\end{tabular}
\end{center}
\caption{Hyperparameters for our experiments. FF is short for feed-forward layer.  }
\label{hyperparameters}
\end{table}

\subsection{Hyperparameters}
The hyperparameters for our experiments are shown in Table \ref{hyperparameters}. For De-En, we follow the setting of Transformer-small. For En-De, we follow the setting of Transformer-base.

\subsection{Training}
All our models are trained on one CPU (Intel i7-5960X) and one nVidia 1080Ti GPU. The implementation of model is based on fairseq-0.6.2. We choose Adam optimizer with $\beta_1=0.9$, $\beta_2=0.98$, $\epsilon=10^{-9}$ and the learning rate setting strategy, which are all the same as \cite{DBLP:conf/nips/VaswaniSPUJGKP17},

$lr$ = $d^{-0.5}$ $\cdot$ min($step^{-0.5}$, $step$ $\cdot$ $warmup_{step}^{-1.5}$)

\noindent where $d$ is the dimension of embeddings, $step$ is the step number of training and $warmup_{step}$ is the step number of warmup. When the number of step is smaller than the step of warmup, the learning rate increases linearly and then decreases. 

We use beam search decoder for De-En task with beam width 6. For En-De, following \cite{DBLP:conf/nips/VaswaniSPUJGKP17}, the width for beam search is $6$ and the length penalty $\alpha$ is 0.2. The batch size is 1024 for De-En and 4096 for En-De. We evaluate the translation results by using multiBLEU.

\begin{table}[htb]
\small
\newcommand{\tabincell}[2]{\begin{tabular}{@{}#1@{}}#2\end{tabular}}

\begin{center}
\begin{tabular}{lcccc}
\toprule
\multirow{2}{*}{\vspace{-2mm}Model} &    \multicolumn{2}{c}{De-En}&\multicolumn{2}{c}{En-De} \\
\cmidrule{2-5}

&  BLEU & \#Para & BLEU & \#Para \\
\hline
TF (small) &  36.5 & 42M &- &- \\
TF (base) &  -&-& 27.1 &66M\\

\hline

\tabincell{c}{Our model} & 37.1 &50M & 27.5 &77M\\
\hdashline
 +half-dim & 37.5& 47M& 27.4 & 71M\\
 +gate & 37.3&57M & 28.0 & 80M\\
 +self-gate &36.9&53M& 27.6 & 77M\\
 +shared-qkv\&gate &37.1&51M&27.7&75M\\
 +half-dim\&gate  & 37.2 & 50M& 28.2 & 74M\\
 +half-dim\&gate  &37.5&47M&27.7&70M\\
 \hspace{2mm}{\&shared-qkv}&&&&\\
\hline

\bottomrule

\end{tabular}
\end{center}
\caption{BLEU scores on De-En and En-De. The baselines for De-En task and En-De task are the Transformer-small and the Transformer-base, respectively. We use multi-BLEU for De-En and En-De.} 
\label{bleu1}
\end{table}

\subsection{Results}
\label{results}
Our baselines for En-De and De-En are Transformer-base and Transformer-small. Without methods to enhance the performance, our model is a variant of the Transformer with three parts of self-attention. We test several methods such as half-dimension (half-dim), weight-gate (gate), shared-query-key-value (shared-qkv) and self-gate (self-gate).
\begin{figure}
   
    \includegraphics[scale=0.53]{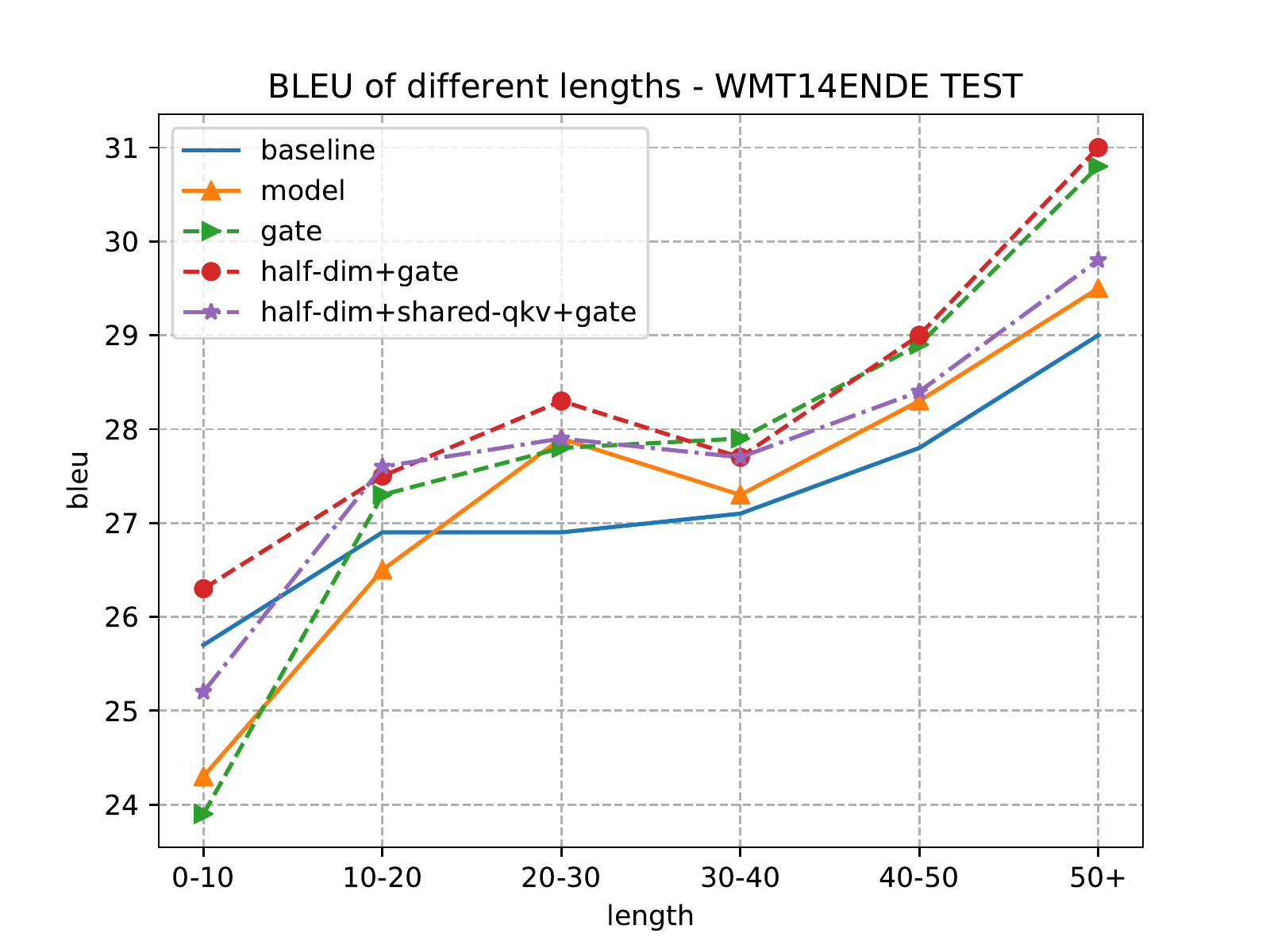}
    \caption{BLEU points of different lengths.}
    \label{fig:bleulengths}
\end{figure}

Table \ref{bleu1}  compares our methods with the original Transformer, showing that our method enhances all the performance in De-En and En-De tasks and outperforms all baselines. For De-En tasks, our model with half-dimension gets the best performance of 37.5 BLEU points. For En-De tasks, combination of method is half-dimension and weight-gate gets the best performance of 28.2 BLEU points outperforming the Transformer-base by 1.1 BLEU points. Different combination of our methods are tested in our experiment. Our model with half-dimension gets the best performance with 47 million parameters in De-En. Our model with half-dimension and weight-gate gets the best performance with 74 million parameters in En-De tasks.

To compare method of fusion, calculating the sum of representation gets the lowest performance of 27.5 BLEU points. Using weight-gate to weight different groups of subgraphs is most effective method. 

Figure \ref{fig:bleulengths} shows the relationship between performance and length of input sentences. With the increase of length of input sentences, all our methods outperform baseline.

\begin{figure}
    
    \includegraphics[scale=0.53]{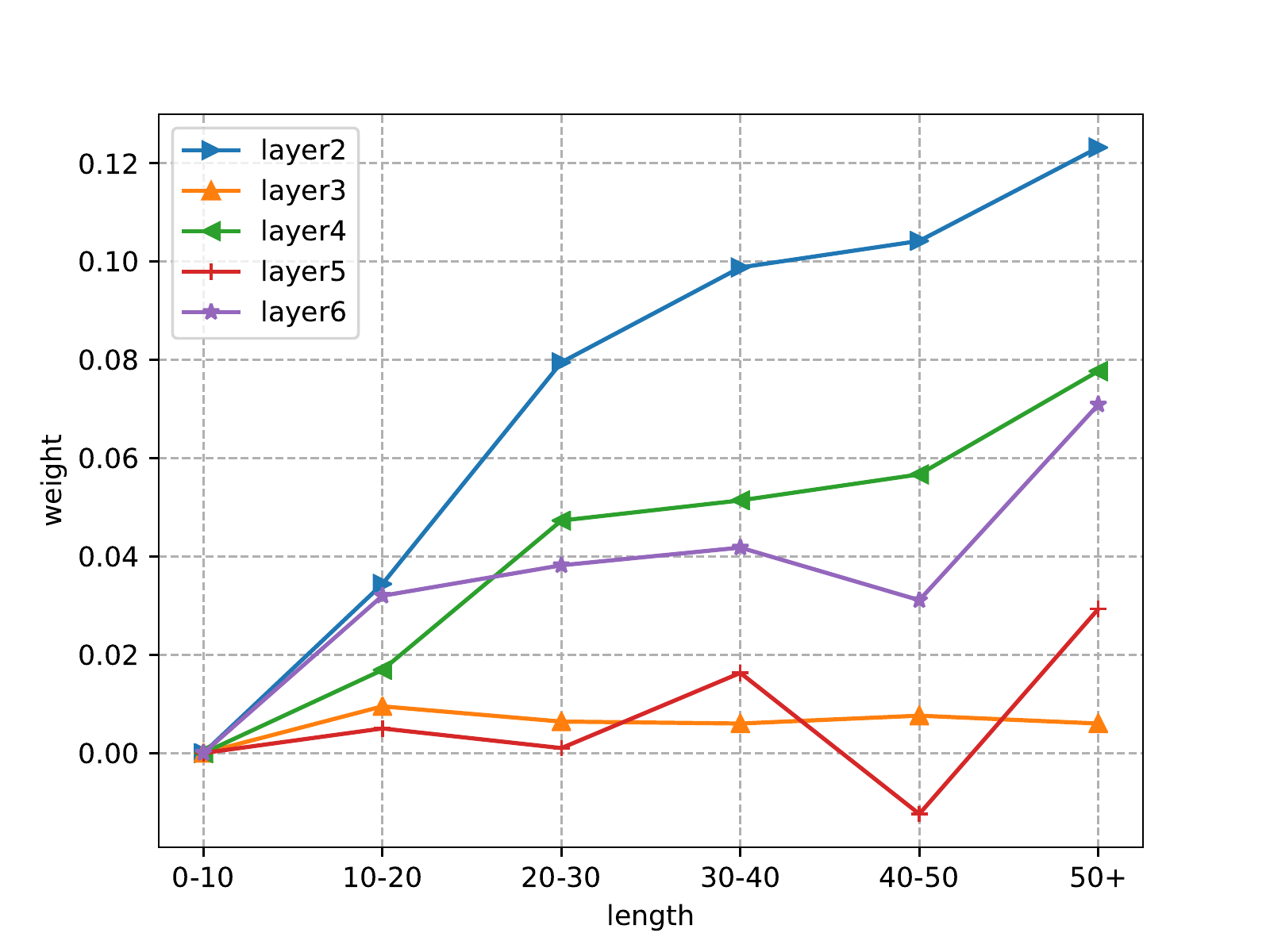}
    \caption{Change of weight of subgraphs of high order with increase of sentence length in different layers.}
    \label{fig:gateslength}
\end{figure}

Figure \ref{fig:gateslength} shows that how weight of subgraph of high order change in different layers with increase of sentence. This result is based on combination of half-dimension and gate. A weight in Figure \ref{fig:gateslength} is actually the difference between weights of different lengths and the weight when length of input sentence is from 0 to 10. The trend of weight change means that when the length of sentence increases, our model will pay more attention on subgraphs of higher order.

\begin{figure}
\includegraphics[scale=0.53]{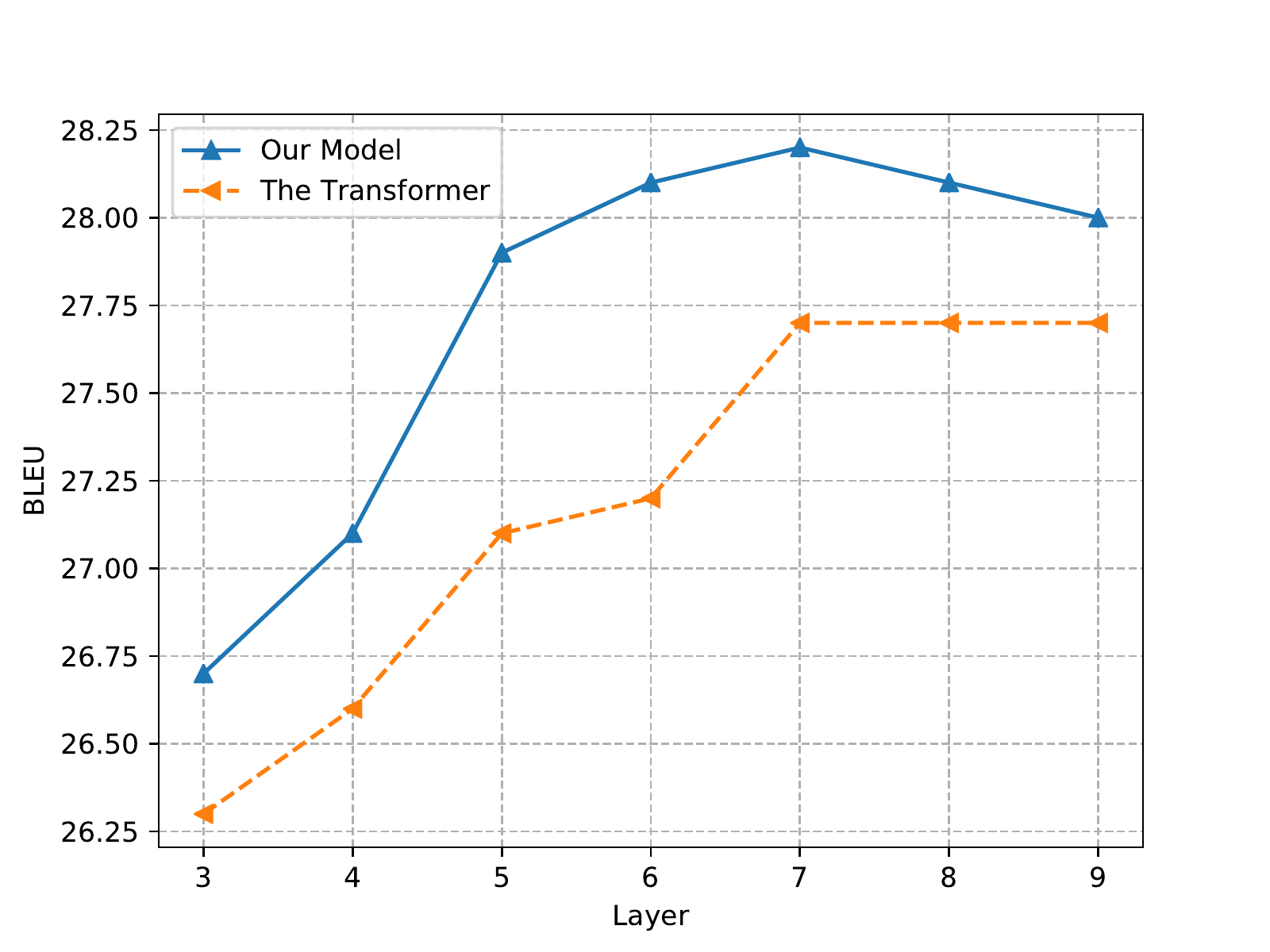}
\caption{BLEU points of our model and the original Transformer with different layers on En-De.}
\label{fig:layersbaseline}
\end{figure}

Figure \ref{fig:layersbaseline} shows that how BLEU point of our model (half-dim+gate) and the original Transformer change with different layers in model. With the increase of number of layers, our model outperforms the original Transformer and gets the best performance with 7 layers.

\section{Analysis of Result}
\subsection{Model}

Our model can only outperform baseline 0.4 BLEU points in En-De task before weighting subgraph, and outperform baseline more than 0.9 BLEU points after weighting subgraph. It means that model cannot find out important representation even if we group them. It also proves that the original Transformer cannot distinguish subgraphs of different orders. In some sense, highway connection between different layers, such as work in \cite{dou-etal-2018-exploiting, wang-etal-2018-multi-layer}, is similar as our work. These works weight representations of different layers and make model focus on representation of some layers. The difference between our work and them is the granularity selected to process.
\subsection{Dimension of Models}
Table \ref{bleu1} shows that model with half dimension can get a similar or better result compared with model with full dimension. Larger dimension gives model a larger vector to contain information and produce more features. Results of our model does not mean larger dimension is unimportant. Though we use less parameters, our model can capture subgraph more clearly. Our model can distinguish subgraph with different orders with three independent  parts of self-attention. 

Besides, model is more difficult to be trained with more parameters and easier to be overfitting. Drop of parameters can produce the effect of dropout and avoid overfitting of model. It makes model with half dimension better than model with original dimension.

\subsection{Fusion Methods}
Table \ref{bleu1} compares methods of representation fusion and weight-gate get the best performance. Method of self-gate is not the best method as expected.

Calculating sum of representation performs worst because this method ignores importance of different subgraph and fails to weight representations. It makes model cannot distinguish different subgraphs, which is same as the original Transformer. Model with this method cannot take advantage of grouping subgraph of different orders.

Compared with using weight-gate to weight groups, self-gate is much better to weight every group of subgraphs which is impossible for one gate to finish. Self-gate can consider relationship between every pair of representation and find out the most important one. 

However, as we discussed above, one layer of the Transformer can extend the set of subgraph and capture information of input. Self-attention is the key to achieve it. Using self-attention to weight representations may capture information of subgraph of higher order and generate unnecessary redundancy. It will disrupt the original generation order of subgraph and bring noise to model. Besides, this method makes model deeper which makes model difficult to train.

Although method of weight-gate cannot distinguish every kind of representation, it makes model focus on parts of them. In fact, using the same query, key and value to produce representations, there are some stable relationship between them. Dividing them into two groups can maximize this relationship and allow model to capture it.

\section{Related Work}
Several variants have been proposed to improve performance of the original SAN-based model.  \citet{DBLP:conf/naacl/ShawUV18} proposed relative position representations in the self-attention mechanism to replace the absolute position encoding and it enhances the ability of capturing local information of the input sentence. \citet{DBLP:conf/nips/HeTXHQ0L18} shared the parameters of each layer between the encoder and decoder to coordinate the learning between encoder and decoder. BERT \cite{DBLP:conf/naacl/DevlinCLT19} is a language model which is  to pre-train deep bidirectional representations from unlabeled text by jointly conditioning on both left and right context in all layers.  \citet{DBLP:conf/acl/DaiYYCLS19} enabled the Transformer to learn dependency beyond a fixed length without disrupting temporal coherence. \citet{DBLP:conf/naacl/Koncel-Kedziorski19} propose a Graph-based model on text generation. 
\citet{DBLP:conf/emnlp/ZhuLZQZZ19} use graph structures for AMR. \citet{DBLP:conf/aaai/CaiL20} propose a graph structure network for AMR.

\section{Conclusions}

Instead of treating MT as seq2seq learning in the current NMT, this work presents the first graph-to-sequence NMT model, Graph-Transformer. Considering that graph other than sequence is a generalized structure formalism, modeling graph information inside model may facilitate NMT model to learn important subgraph information from source. As the multigraph defined over the sentence cannot be immediately by one part of the model such as just one layer, we assign every layer of the model to learn subgraphs with different orders, respectively. As our model implementation, we revise the SAN so that it may acquire such explicit subgraph information through our introduced incremental representation. Results of experiments show that our method can effectively boost the Transformer.

\bibliographystyle{aaai21}
\bibliography{aaai2021}
\end{document}